\documentclass{egpubl}
\usepackage{envirvis2023}

\WsPaper         
\BibtexOrBiblatex

\electronicVersion 
\PrintedOrElectronic

\ifpdf \usepackage[pdftex]{graphicx} \pdfcompresslevel=9
\else \usepackage[dvips]{graphicx} \fi
\usepackage{lineno}
\usepackage[T1]{fontenc}
\usepackage{dfadobe}  
\usepackage{t1enc,dfadobe}

\usepackage{egweblnk}

\usepackage{microtype}                 
\usepackage{textcomp}                  
\usepackage{mathptmx}                  
\usepackage{times}                     
\usepackage{cite}                      
\usepackage{tabu}                      
\usepackage{booktabs}                  
\usepackage{caption}
\usepackage{gensymb}
\usepackage{lipsum}

\usepackage[compact]{titlesec}         
\titlespacing{\section}{0pt}{5pt}{0pt} 
\titlespacing{\subsection}{0pt}{1pt}{0pt}
\titlespacing{\subsubsection}{0pt}{1pt}{0pt}
\title{A Hybrid 3D Eddy Detection Technique \\ Based on Sea Surface Height and Velocity Field }

\author[W. Hua, et al.]
      {Weiping Hua$^{1}$
         Karen Bemis$^{1}$ Dujuan Kang$^{1}$ Sedat Ozer$^{2}$ Deborah Silver$^{1}$
       \\
        $^1$Rutgers, The State University of New Jersey, New Brunswick, USA\\
        $^2$ Ozyegin University, Istanbul, Turkey
      }

\teaser{
 \centering
\includegraphics[width=.95\textwidth]{pictures/more_datasets.jpg}
\caption{The detected eddies (yellow to blue) found by applying our proposed sea surface height and velocity field based hybrid eddy detection algorithm are shown for two separate ocean simulations. For context, we include the seafloor (blue-grey), land (brown) and the major ocean currents (red). Eddy colors represent net velocity magnitude, faster on surface (yellow) and slower below (blue). Eddy boundaries represent the extent of coherent eddy rotation rather than a particular value of velocity or vorticity. Short red bars in the center of each eddy show its recent movement. Ocean currents are determined as 25\% of the maximum velocity magnitude in each the respective datasets.  (a) shows the results on a simulation of the North Atlantic Ocean and (b) shows the results on a simulation of the North Pacific Ocean. }
    \label{fig:more dataset}
}

\begin{document}
\maketitle

\begin{abstract}
Eddy detection is a critical task for ocean scientists to understand and analyze ocean circulation. In this paper, we introduce a hybrid eddy detection approach that combines sea surface height (SSH) and velocity fields with geometric criteria defining eddy behavior. Our approach searches for SSH minima and maxima, which oceanographers expect to find at the center of eddies. Geometric criteria are used to verify expected velocity field properties, such as net rotation and symmetry, by tracing velocity components along a circular path surrounding each eddy center. Progressive searches outward and into deeper layers yield each eddy's 3D region of influence. Isolation of each eddy structure from the dataset, using it's cylindrical footprint, facilitates visualization of internal eddy structures using horizontal velocity, vertical velocity, temperature and salinity. A quantitative comparison of Okubo-Weiss vorticity (OW) thresholding, the standard winding angle, and this new SSH-velocity hybrid methods of eddy detection as applied to the Red Sea dataset suggests that detection results are highly dependent on the choices of method, thresholds, and criteria. Our new SSH-velocity hybrid detection approach has the advantages of providing eddy structures with verified rotation properties, 3D visualization of the internal structure of physical properties, and rapid efficient estimations of eddy footprints without calculating streamlines. Our approach combines visualization of internal structure and tracking overall movement to support the study of the transport mechanisms key to understanding the interaction of nutrient distribution and ocean circulation. Our method is applied to three different datasets to showcase the generality of its application.Our code is available at 
\url{https://github.com/VizlabRutgers/Hybrid-Eddy-detection}

\end{abstract}




\section{Introduction}\label{introduction of the paper}

Accurate detection of eddy-like structures in ocean simulations is an essential step for understanding and analyzing the dynamics in ocean simulations. An eddy is a spiral-like spinning structure akin to a vortex or vortex line in 3D space; as such, it is expected to have a roughly circular \cite{robinson1991coherent} or elliptical shape \cite{chen2019intrinsic} on the 2D plane, to have a velocity structure reflecting a coherent, spatially consistent rotation, and to have a minima or maxima in the sea level surface \cite{chelton2011global}. Eddies play an important role in ocean dynamics as agents of vertical and horizontal transportation of salinity, temperature and nutrients \cite{klein2009oceanic}. They have a profound influence on ocean biology, ecology and biogeochemistry \cite{della2019overview}. To analyze the influence of eddies, a crucial task is detecting the eddy structures within a dataset. Many approaches have been proposed based on 2D satellite datasets \cite{nencioli2010vector,kang2013gulf,qiu2019deformation}. Traditionally, eddy detection methods can be divided into three categories \cite{nencioli2010vector}: value-based, geometry-based, and hybrid. In our earlier work, we applied an approach solely based on thresholding the Okubo-Weiss (OW) parameter \cite{liu2019visualizing} to detect eddy structures efficiently for the SciVis Contest 2020. However, the hybrid approaches are frequently preferred for eddy detection as in \cite{chaigneau2008mesoscale,yi2014enhancing}, because they typically consider more attributes, such as such as streamlines, sea surface height (SSH) and velocity field, in describing an eddy. Therefore, to improve on the detection accuracy of our previous approach in \cite{liu2019visualizing}, we introduce a novel hybrid approach.  

In our new hybrid approach, instead of relying on the computation of streamlines, we use SSH information, velocity field and velocity magnitude combined with specific geometric tests. Our approach first focuses on detecting the center of the eddy and then grows the eddy around that center until a stopping criterion is met on each plane. Since we do not compute the streamline, our computational time is lower than the geometry \cite{sadarjoen2000detection} techniques. Our experimental results visually compare the detected eddies from our proposed approach to both a winding-angle method (geometry-based approach) and to an OW-based method (value-based approach) on the Red Sea dataset \cite{toye2017ensemble}. Our method is applied to two additional datasets to showcase the generality of its application. Our visualizations and experiments highlight that our proposed approach provides eddy structures with verified rotation properties and 3D visualization of the internal structure of eddies. Furthermore, our domain scientists have stated that the approximated eddy boundaries extracted by our approach are more meaningful and complete compared to the other approaches. Finally, our results are visualized and tracked across time frames to support the study of the transport mechanisms key to understanding the interaction of nutrient distribution and ocean circulation.

\section{Related work}
Overall, eddy detection methods can be divided into three categories: value-based, geometry-based, and hybrid methods \cite{nencioli2010vector}. The value-based methods typically rely on thresholding one or two physical parameters. A commonly used category of value-based methods uses the Okubo-Weiss (OW) parameter \cite{okubo1970horizontal,weiss1991dynamics,raith2021uncertainty,zhang2022peviz}, a measure of the balance between deformation and rotation. OW-based eddy detection algorithms first calculate the OW parameter, then extract the eddy structure with a user-assigned threshold. However, some researchers \cite{chaigneau2008mesoscale} reported that the OW-based approaches would identify regions which are not eddies. Another work in \cite{rave2021multifaceted} uses several parameters to extract ensemble-averaging eddy structures with varying probabilities. The geometry-based methods analyze the geometric structure of the flow field and the related streamlines to obtain the structure of the eddy, but these methods can be computationally expensive. A typical geometry method is the winding-angle (WA) approach proposed in \cite{sadarjoen2000detection}. This approach calculates the cumulative angle change along a streamline to determine if a closed streamline is the boundary of an eddy-like structure. A recent study applied \cite{friederici2021winding} this method to the simulation of the Red Sea dataset and visualized the properties across the full ensemble. Another kind of geometry method, the vector geometry method \cite{nencioli2010vector}, first extracts the eddy center and then extracts the eddy boundary based on the maximum current speed around the center. The hybrid approaches combine elements of both the value-based and geometry-based approaches to achieve the ease of boundary detection of value-based algorithms while computing faster than the geometry-based approaches. Several works \cite{chaigneau2008mesoscale,yi2014enhancing,matsuoka2016new} use the sea surface height (SSH) or sea level anomalies (SLA) to detect eddy centers but vary in how they extract a boundary. The work in \cite{chaigneau2008mesoscale} draws the boundary of each eddy from the streamline, which is computed on the 2D plane. Another work \cite{yi2014enhancing} uses the OW to extract the boundary instead of computing the streamline. This OW based approach which relies on a SLA-defined eddy center gives a more precise boundary but usually requires a fixed and predetermined threshold. Finally, \cite{matsuoka2016new} limit detections to SSH extrema coincident with OW minima, define an inner boundary based on the curvature of the SSH, and define an outer boundary based on identifying a high speed region stream around the eddy using a fixed threshold on velocity magnitude.  Our new method proposed herein has some similarities in combining SSH and velocity data but does not rely on fixed thresholds.

Some recent literature focuses on different techniques for eddy detection. Some researchers \cite{xie2020modeling} use a neural network to detect eddies, which requires training data to be prepared manually or by another method which is costly. Some researchers also use a statistical volume rendering framework for 3D visualization of eddy simulation ensembles\cite{athawale2021statistical,friederici2021winding}.

\begin{figure}[tb]
 \centering 
 \includegraphics[width=0.8\columnwidth]{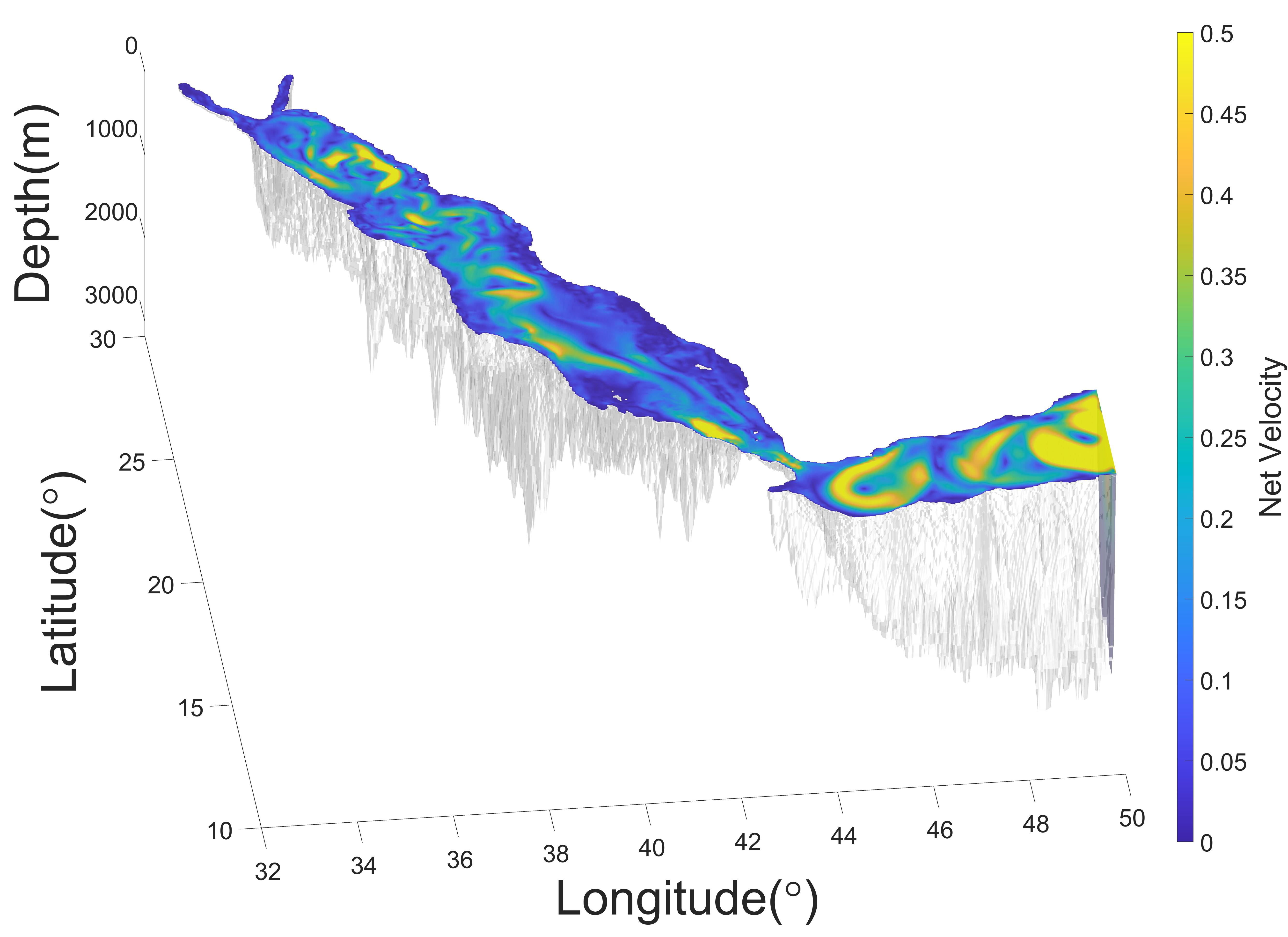}
 \caption{Visualization of the original Red Sea dataset. The color indicates the net velocity on the surface.}
 \label{fig:full dataset}
\end{figure}

Our primary examples use an ocean simulation of the Red Sea provided in the SciVis 2020 contest\footnote{\url{https://kaust-vislab.github.io/SciVis2020/results.html}} by the Red Sea Modeling and Prediction Group \cite{toye2017ensemble}. This rectilinear dataset includes 500 x 500 x 50 voxels (3D grid cells) of 0.04°×0.04° (4km x 4km) resolution on each layer and 50 vertical layers, varying from 4m thick on the surface to 300m near the bottom. The dataset includes 60 frames (or time steps) in each ensemble member covering one month of simulation time and 50 ensemble members in total. Figure \ref{fig:full dataset} visualizes the first frame of ensemble member 1 from the Red Sea dataset.

\section{Motivation} \label{Motivation of new approach}

\begin{figure}[tb]
    \centering
    \includegraphics[width=0.8\columnwidth]{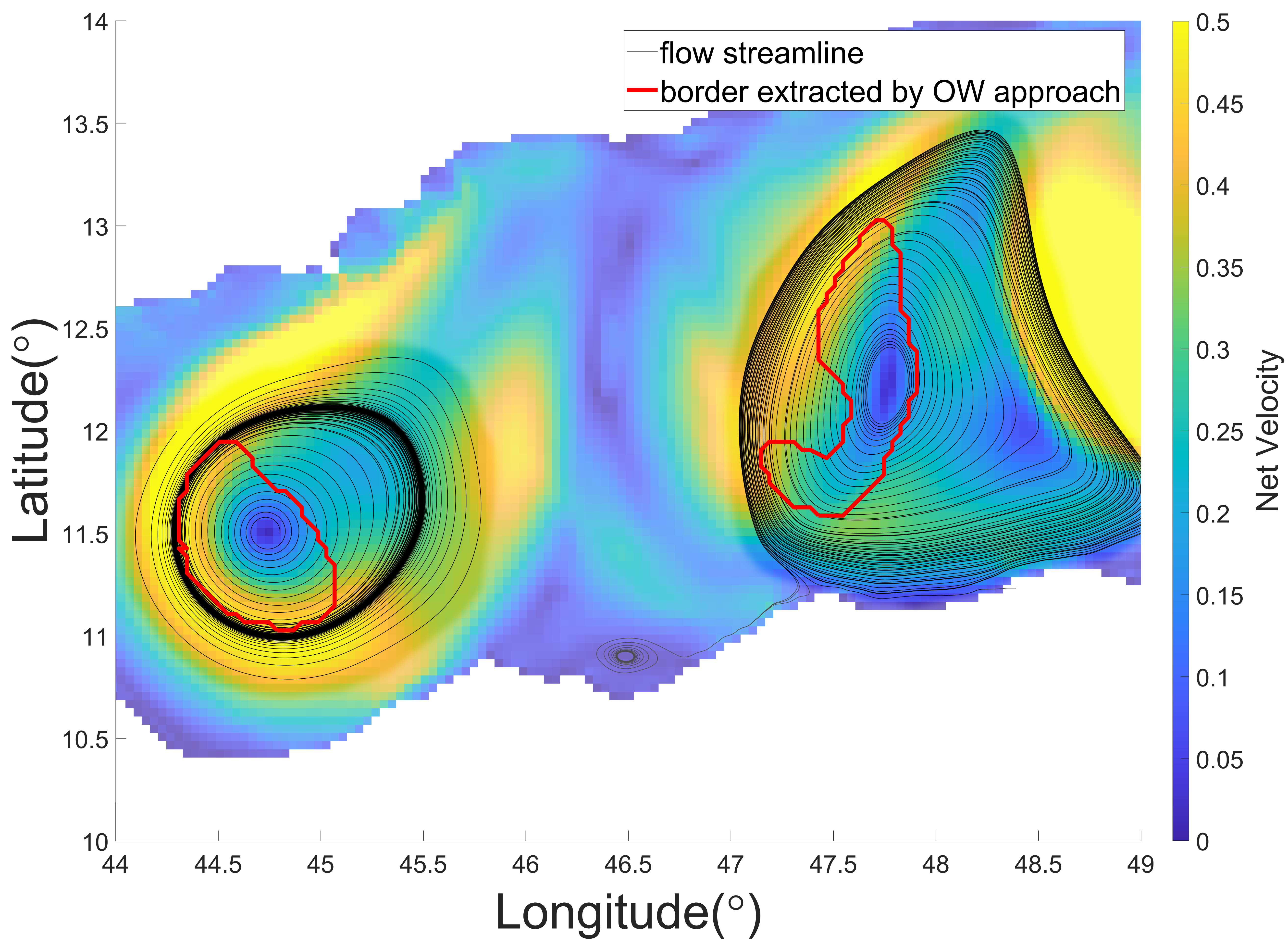}
    \caption{Two eddy borders are extracted by the OW based approach. The left border belongs to Eddy 15 and the right border belongs to Eddy 18 from the Red Sea dataset.}
    \label{fig:Two OW example}
\end{figure}

\begin{figure}[tb]
 \centering 
 \includegraphics[width=.75\columnwidth]{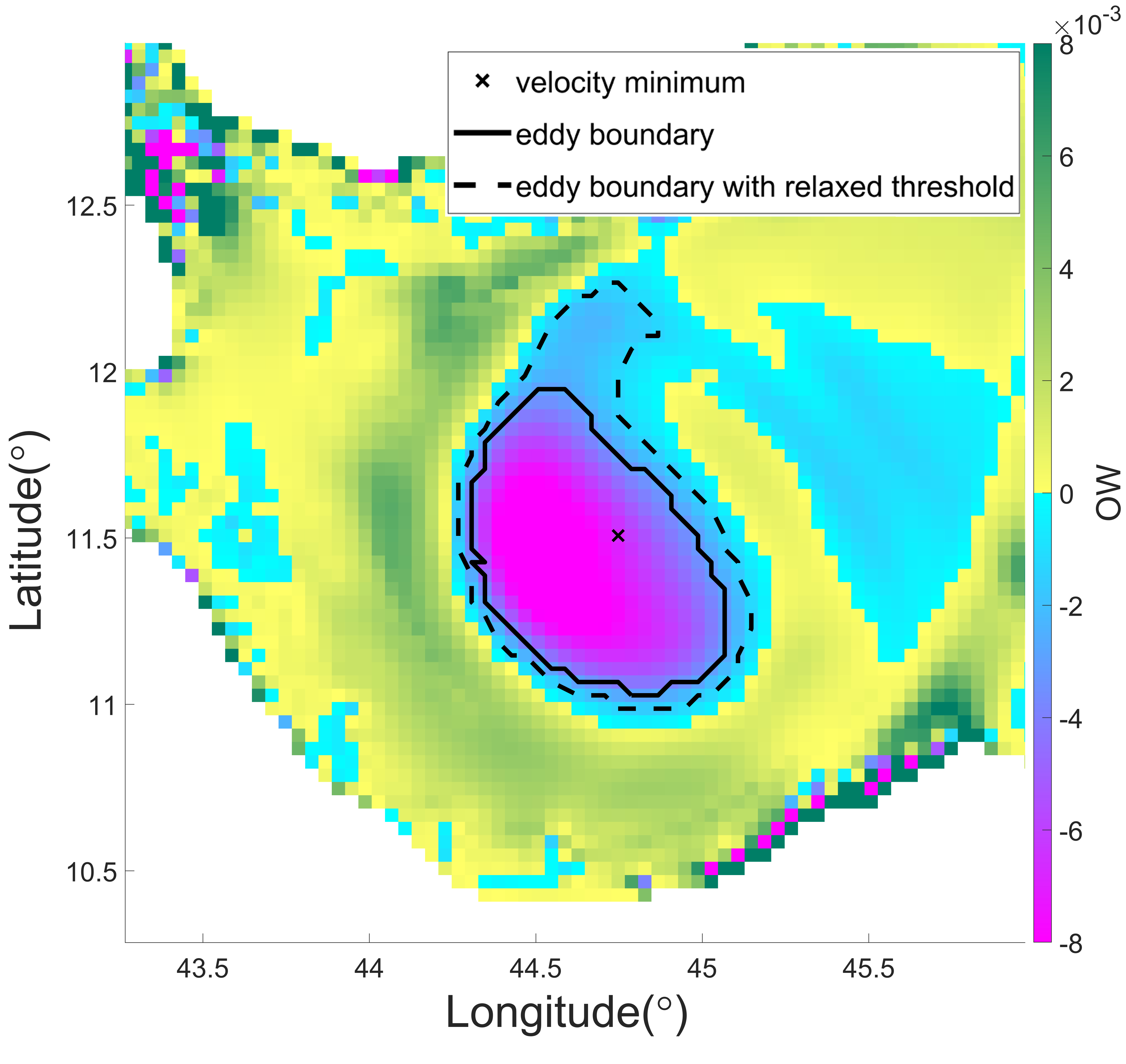}
 \caption{Visualization of the border of the Eddy 15's by using two approaches:(i) by using the original OW based approach and (ii) by using the relaxed OW based approach.}
 \label{fig:previous detail}
\end{figure}

The new method proposed in this study is motivated by an assessment of results from a previous study using a value-based approach combining a universal feature extraction implementation \cite{silver1996volume}\cite{ozer2013activity} and dynamic thresholding of the OW parameter. In that work, we first located the local minimum of OW, then obtained the dynamic threshold from the product of the local minimum of OW (typically a negative value for eddies) with a predefined constant. The eddy is extracted by the region growing algorithm started from the local minimum of OW with this dynamic threshold. After reanalyzing our work therein, we observed that the extracted structures had complex shapes, whereas the eddy is expected to have a roughly circular or elliptical shape \cite{matsuoka2016new}. Figure \ref{fig:Two OW example} shows two examples  where a) the extracted region is much smaller than the closed streamline regions, encompassing only the core of the eddy and b) the boundary of the extracted region is irregular and non-conforming with the shape of the streamlines. Figure \ref{fig:previous detail} shows that relaxing the threshold towards zero produces only a minor expansion of the extracted boundary and still fails to conform to the streamlines. As oceanographic definitions of eddies imply that the boundary of an eddy should be roughly a closed streamline, such examples raise concerns about the accuracy and completeness of OW-based approaches.

Additionally, we initially assumed, based on how oceanographers describe eddies, that the local minimum of OW lies precisely at the center of the eddy and our algorithm was solely based on the location of that local minima. However, in some cases, the local minimum of OW is significantly offset from the local minima of the net velocity. Figure \ref{fig:previous detail} demonstrates that this offset  can result in an extremely asymmetric shape. Based on these observations, we decided to consider other approaches to detect eddies. The streamline method should detect a better border of the eddy 
given that it is the best ground-truth for eddy identification but the accurate computation of streamlines is computationally expensive and typically focused on 2D surface identification of eddies (note the exception of \cite{friederici2021winding}). Thus, we have pursued an hybrid method to take advantage of the best of value-based and geometry-based approaches. 


\section{A new hybrid-based approach}

Our proposed hybrid eddy detection method has three main steps: first it locates the potential centers of eddies using SSH and net velocity, then it verifies the rotational properties of the centers, and it finally extracts the whole structure. In this section, we will describe our approach step by step.

\subsection{Locate the candidate eddy centers}
Many previous researchers have shown the benefits of locating the centers of eddies first and then assessing for the boundary and other characteristics as in \cite{kang2013gulf}. The center of an eddy is a unique and conspicuous property of the geometric eddy structure. A cyclonic or anticyclonic eddy should have a local maximum or minimum of the SSH at the center of the eddy in the northern hemisphere, respectively \cite{stewart2008introduction}. Finding those SSH extrema from the original dataset on the sea surface layer is the first step in our algorithm.

In an ideal axisymmetric eddy, the local extremum of SSH will coincide with the local minima of net velocity and OW as well as being the visual critical point in the neighborhood of the velocity field. Traditionally, researchers will consider that critical point as the true center of the eddy. However, our observations have shown that the extrema of SSH are not always coincident with eddy centers (see Figure \ref{fig:Two OW example} and \ref{fig:previous detail}). Therefore, we consider the local minimum of net velocity as a secondary criterion to detect the center of an eddy. We first search for the local extrema of SSH by a sliding square window with the width of \textbf{$Re$}. Then, we can search for the local minimum of net velocity by another sliding square window with the width of \textbf{$Rv$} around the neighborhood of each SSH extremum. The local minima of net velocity detected are regarded as eddy center candidates. However, other flow structures besides eddies, such as meanders, may also have local minima of velocity, which could result in false detection of eddy center candidates. Thus, a verification on each detected center is necessary before assessing the border of eddy.

\subsection{Eddy center verification} \label{verify centers}

Here, we check if the eddy center candidates actually belong to a real eddy. Along with many researchers, we assume that an eddy has a ellipse or a circular shape in general \cite{chen2019intrinsic}. We note that an ellipse might best describe the overall shape of an eddy, but even that will miss the details of eddy shape due to the complexity of the background flow (see  Figure \ref{fig:Two OW example}). Thus, we choose to inspect the velocity vector along a pre-specified circular path around the center candidate without requiring expensive computation on the geometric structure of each eddy. Our pre-specified circular path starts with a minimum radius of three pixels (that is, grid cells) around the center and is checked in the counterclockwise direction starting at the bottom-most point.

\begin{figure}[tb]
    \centering
    \includegraphics[width=0.8\columnwidth]{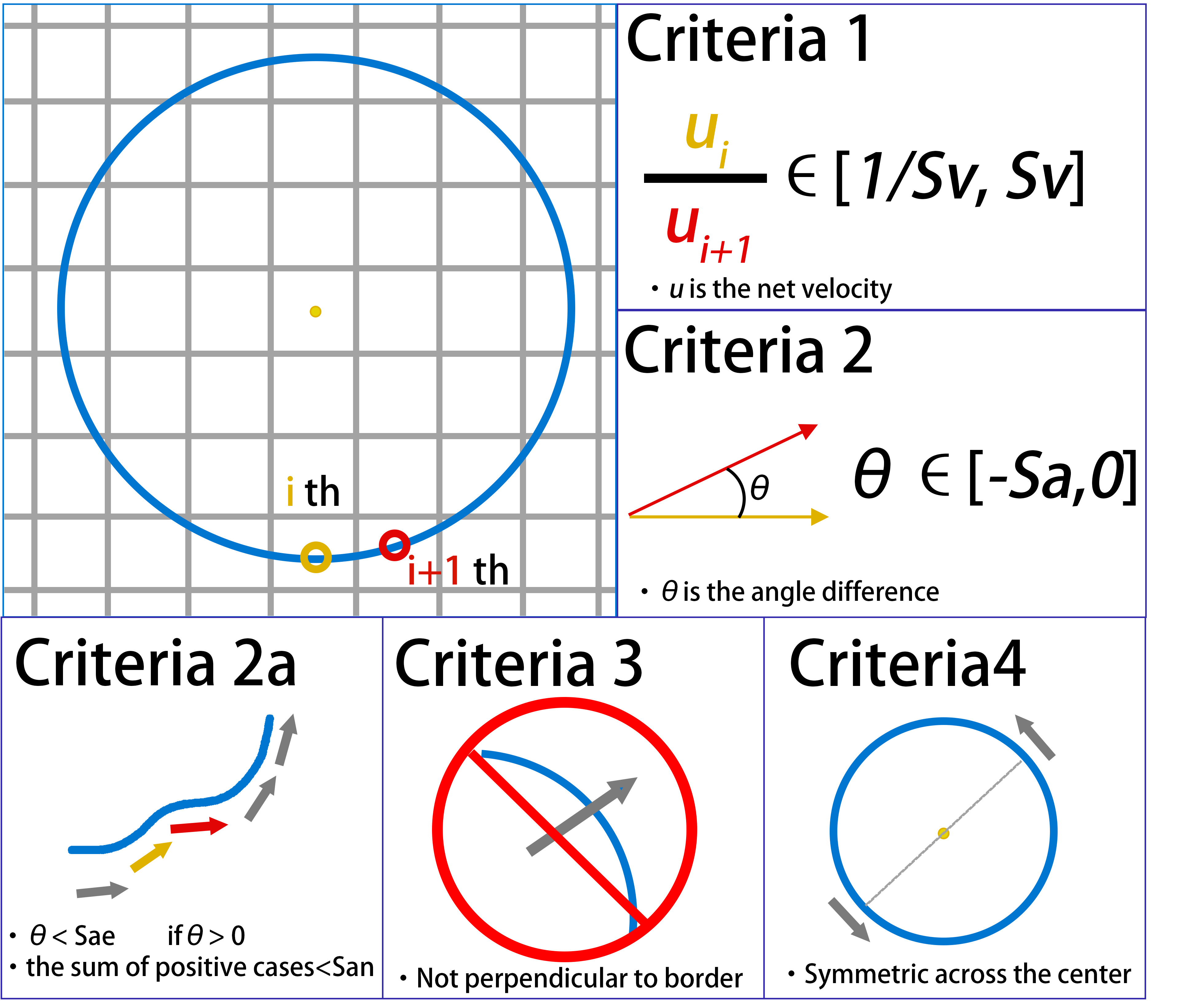}
    \caption{A visual summary of the eddy center verification conditions. $U$ represents the scalar net velocity along longitude and latitude, respectively. $\theta$ represents the angle between two consecutive velocity components. Direction of analysis is counterclockwise from the bottom-most point. The red and yellow circles indicate a pair of consecutive velocity vectors along the border which are being verified.}
    \label{fig: criterions}
\end{figure}

In this paper, four criteria based on the circular inspection are proposed (shown in figure \ref{fig: criterions}) to qualify a candidate center as an actual eddy center. Each criteria considers a constraint to eliminate false eddy detection. Criteria (1) and (2) require the rotation action being consistent along the boundary while allowing some minor anomalies. Criteria (3) checks to make sure the eddy structure has a reasonably circular rotation trace and Criteria (4) forces the direction of each point to be symmetric across the center.

\textbf{Criteria 1:} The two consecutive points on the border should have similar net velocities. We use a threshold, $Sv$, to constrain the ratio of net velocities (Euclidean norm) of two consecutive points along the border. The ratio of the net velocities $(u(i)/u(i+1))$ must be within the range of ([$1/Sv$, $Sv$]). The reason for using the ratio of the net velocities instead of using a magnitude here is that there might be a large variance on the absolute values of the net velocities between different eddies.

\textbf{Criteria 2:} Since we are looking for a circular flow shape as an approximation of the eddy border, the direction of the consecutive velocity vectors on the border are expected to change minimally and slowly, similar to the patterns described by previous researchers \cite{nencioli2010vector}. As we are selecting points in a counterclockwise traverse, the angular difference between velocity direction at the two consecutive points is expected to be negative and is thus constrained to lie in the interval of $[-Sa, 0]$, where $Sa$ is a predefined non-negative threshold.

\textbf{Criteria 2a:} The background flow or turbulence might cause a sudden change or reversal of the flow around the border. Therefore, it is necessary to allow some exceptions to Criteria 2 so that we can consider such cases. Here we define another parameter $Sae$ as the maximum allowed positive angular difference (representing a small deviation in the direction of flow change). If the angular difference is positive and exceeds the value of $Sae$, the eddy candidate will be immediately rejected. We also limit the number of small deviations allowed to be less than $San$, even if they never exceed the $Sae$ value.   

\textbf{Criteria 3:} The velocity components on the border should be approximately parallel to the tangent direction  ($TD$) of the boundary. So we constrain the velocity component to be in the range of [$TD - Sd$, $TD + Sd$] with a predefined angle threshold $Sd$. We expect the angle between the velocity field and the tangent to be roughly zero and set $Sd$ to 24°, at each quadrant.

\textbf{Criteria 4:} The velocity components are expected to be symmetric across the center. Symmetry is determined by estimating the angle between opposite points on the test circle. We constrain the symmetry angle ($SA$) obtained from symmetric points to lie in the range of [$\pi-Sy$, $\pi+Sy$] with an predefined angle threshold $Sy$; this range allows for slight variations in flow, especially at the border where interactions with background flow may occur. Due to the complicated influence of background flow, we use a rather loose condition, setting the  $Sy$ as 120°.

\begin{figure*}[!htb]
    \centering
    \includegraphics[width=.7\textwidth]{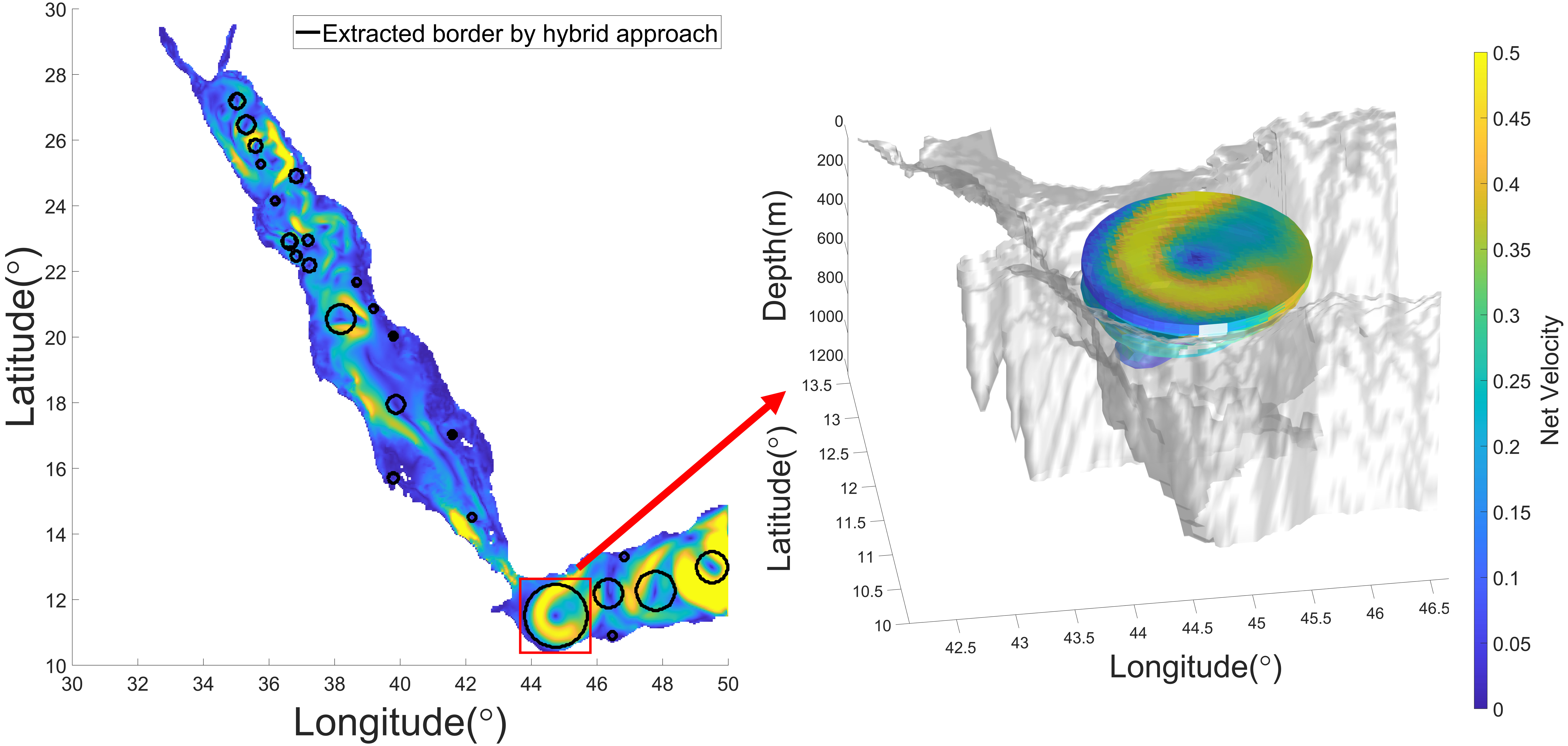}
    \caption{A sample extraction result obtained by our new hybrid method is visualized. The figure on the left visualizes the extraction result on the sea surface; the figure on the right visualizes the 3D border of a particular eddy: eddy 15.}
    \label{fig: new hybrid result}
\end{figure*}

The centers passing these four criteria are considered as true eddy centers. Unlike the traditional geometry-based approach, our hybrid approach does not compute the streamline of the velocity field. Instead, it checks the velocity field on a circle starting at a radius of 3 pixels away from a local minimum of net velocity.  A small initial radius allows us to detect the smallest resolved eddy given the simulation grid spacings. Once the center of the eddy is verified, the boundary of each eddy will be determined based on existing eddy centers.

\subsection{Extract the boundary and the rest structure of the eddy}\label{extract rest structure}

In Section \ref{verify centers} we verified each candidate center as an eddy. With a valid center, we can then extract the rest of the structure of the eddy. As the outer region of the eddy has the same rotation pattern as the core region, the same rotation conditions in section 4.2 can be implemented on a larger radius until it fails. In this way, we can get a circular extraction result of each eddy structure. 

For a deeper layer in the dataset, our oceanography scientists suggest that the center on the next layer would be very close to the one on the current layer. Thus, we don’t need to repeat the search procedure but use the existing center on the current layer as a search seed for a deeper layer and repeat the rest of the process until it touches the bottom of the eddy structure with no deeper valid eddy center. At each plane along the z-axis, these criteria are applied iteratively until the whole structure is extracted.

\section{Experiments}
In our experiments, we use three different ocean simulations including Red Sea \cite{toye2017ensemble}, North Atlantic \footnote{Ocean simulation dataset for North Atlantic region from MOM6 by Dujuan Kang (dk556@envsci.rutgers.edu)} and North Pacific \cite{metzger2017global} datasets. In this section, we present the results of applying our new hybrid approach for detecting eddies to the first frame of ensemble member 1 of the Red Sea dataset. We look at the parameters needed to control the detection process. Finally, we document the relative computational effort needed for each approach. Our new approach detects 26 eddy structures in the first frame of ensemble member 1 in the Red Sea dataset. Figure \ref{fig: new hybrid result} shows the overall detection results with an example of extracted 3D structure by our new hybrid approach.




\subsection{Parameter dependence}\label{parameters}
We use several parameters and criteria in our new approach. In the following, we analyze the impact of parameters on detection and explain our choices for values of those parameters.

\subsubsection{Parameters for locating eddy center candidates}
Four parameters are used to locate the eddy center candidates. \textbf{Parameter $Re$} defines the search box size for the local extrema of SSH. A smaller $Re$ leads to more local extrema of SSH (see black line in Figure 6). Ideally, more local extrema could give more potential structures; in practice, most of those extrema from a small searching region are essentially not eddies, but rather meaningless noise. After discussing with our domain scientists, we choose the search box size as 7 pixels, which could give us most SSH extrema corresponding to eddies but also avoid noise. 

\begin{figure}[tb]
    \centering
    \includegraphics[width=0.7\columnwidth]{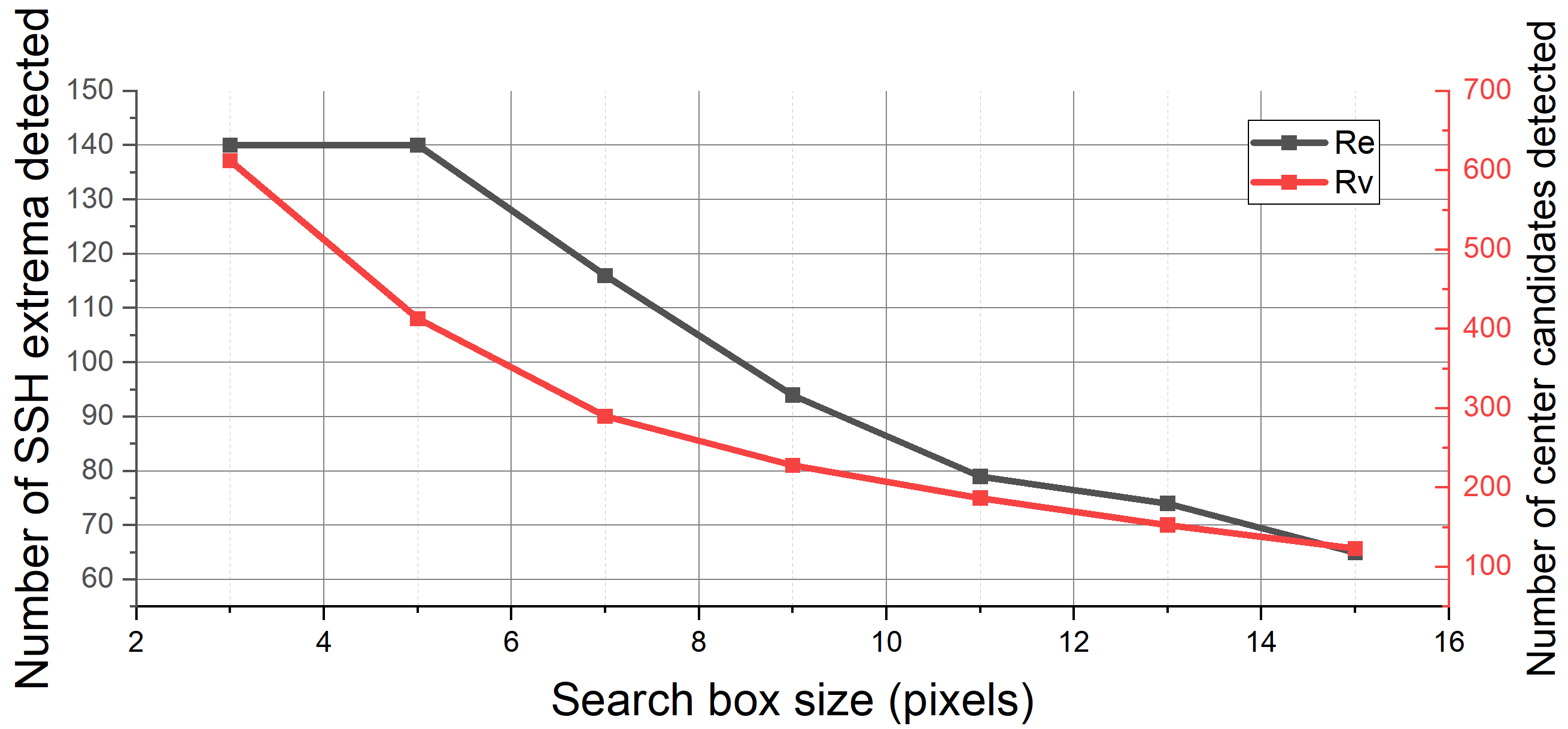}
    \caption{This plot shows that the number of center candidates varies by the size of the search box by using SSH extrema ($Rv$) and nearby velocity minima ($Re$). }
    \label{fig:Search Parameters}
\end{figure}

\textbf{Parameter $Rv$} represents the search box size for velocity minimums around each detected SSH extremum. Small $Rv$ will miss eddy center candidates that are far from the SSH extrema while large $Rv$ will neglect all but one of multiple potential candidates. We explore the results of using different $Rv$ (see red line in Figure \ref{fig:Search Parameters}); the number of detected eddy center candidates peaks when $Rv$ equals 21 pixels. Therefore, we set the $Rv$ to 21 pixels.

\textbf{Parameter $Rc$} is the parameter used to locate the center of eddy on the next layer. The center of eddy in adjacent layers should be very close so we set the $Rc$ to 5 pixels. This seems sufficient to guarantee that every valid center can be found on the next layer.

\textbf{Parameter $Rs$} is the initial radius of each eddy structure. This parameter has no influence on any the eddy structure larger than Rs as such eddies will grow to the same final radius. We set the $Rs$ to 3 pixels since this can give as many candidates as possible. We note that an eddy with radius smaller than 3 pixels may be meaningless in oceanography because of the limited resolution for analysis inside the eddy.

After setting these parameters, we obtain 413 center candidates. Once the eddy candidates are detected, we need to determine which candidates are the centers of eddies.

\subsubsection{Parameters for verifying eddy centers}
Six parameters control the verification of candidates as eddies.  In the results above, our initial setting of parameters yielded 26 eddy centers and 387 rejections. The angular difference criteria (parameters $Sa$, $Sae$ and $San$) account for 78.4\% of the rejections (see summary in Figure \ref{fig: criterions}).  Setting $Sy$ and $Sd$ is straightforward so we'll focus on discussing the other parameters in this section.

\begin{figure}[tb]
    \centering
    \includegraphics[width=0.7\columnwidth]{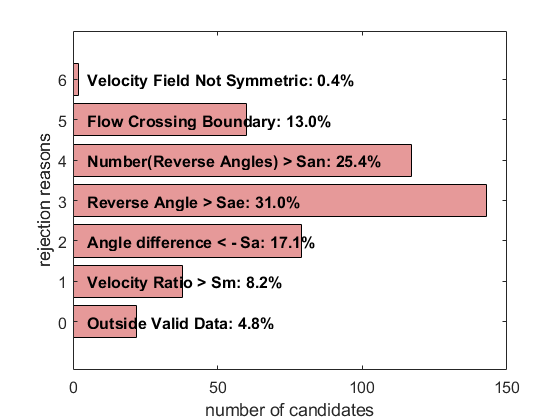}
    \caption{This chart shows the breakdown between the primary reasons that eddy candidates are rejected.}
    \label{fig:Rejection Reason}
\end{figure}

\begin{figure}[tb]
    \centering
    \includegraphics[width=0.6\columnwidth]{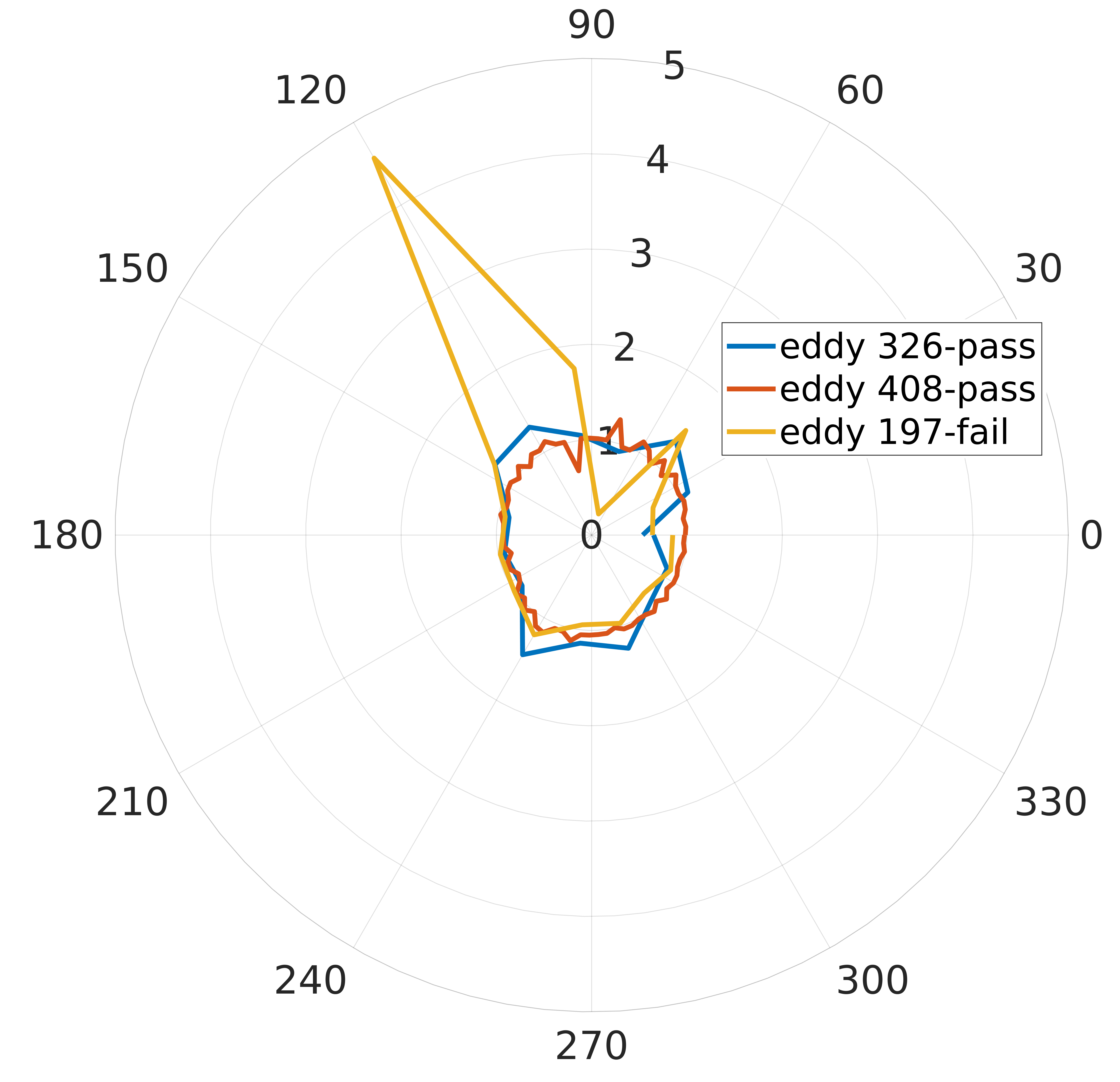}
    \caption{net velocity is plotted as a function of azimuth (which specifies position along the border path) for three example candidates. }
    \label{fig:Sm-1}
\end{figure}

\begin{figure}[tb]
    \centering
    \includegraphics[width=0.8\columnwidth]{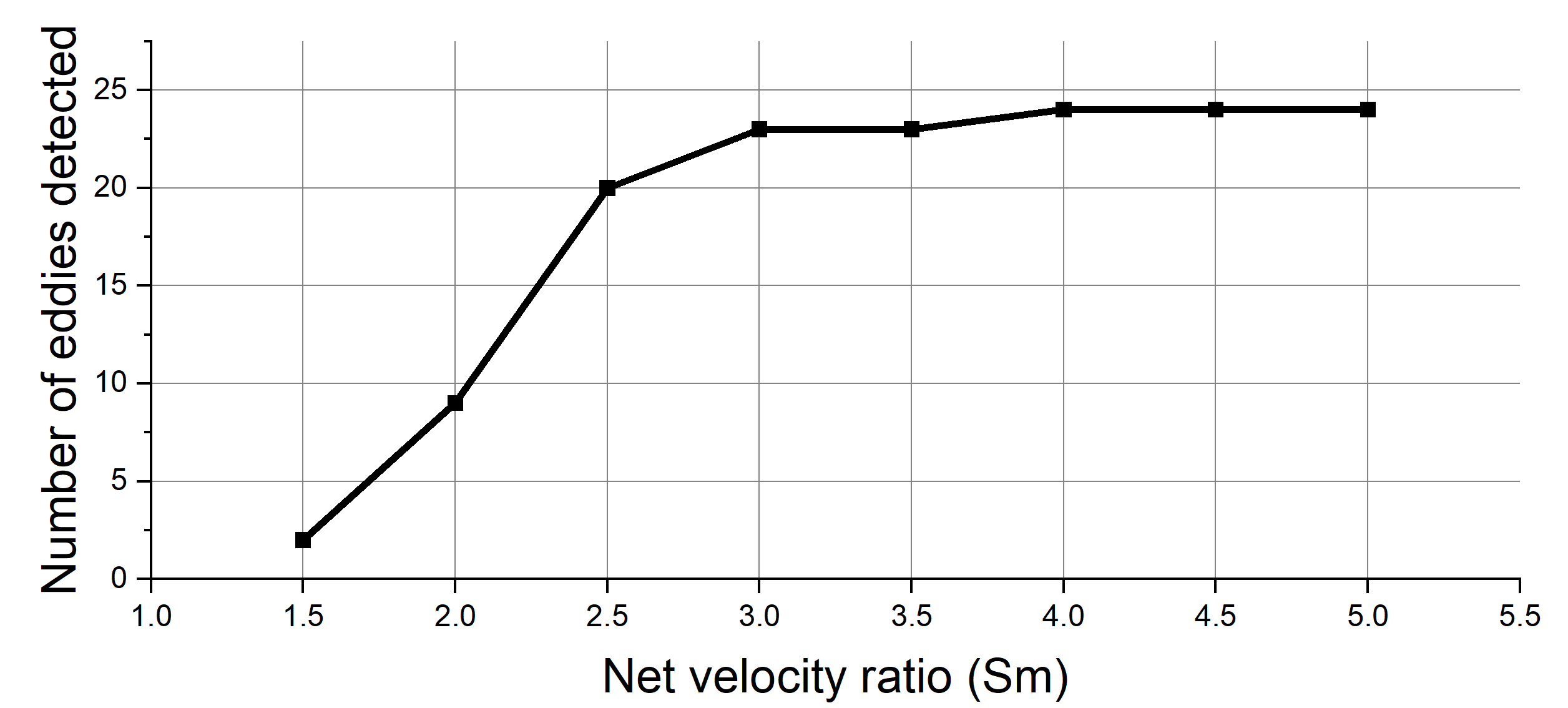}
    \caption{The number of eddies detected changes with the allowed net velocity ratio between consecutive points."}
    \label{fig:condition 1}
\end{figure}

\textbf{Parameter $Sm$} constrains allowable net velocity ratio between consecutive points along the border path to lie in the interval [$1/Sm$, $Sm$]. Figure \ref{fig:Sm-1} shows the variation in net velocity for three example eddies as a function of azimuth along the largest circle tested. For two eddies, the net velocity ratio falls within the interval (but would fail at a larger radius); the third test path shows an eddy candidate at the point of failure. The number of eddies detected changes with $Sm$ in Figure \ref{fig:condition 1} peaking at a value of about 3, so we set $Sm$ to 3.

\textbf{Parameters $Sa$, $Sae$} and \textbf{$San$} constrain the allowable angular difference between two consecutive points. As we traverse each circular path around an eddy candidate in the counterclockwise direction starting at the bottom-most point, the angular difference should be small and negative if the angular flow is roughly circular (see Figure \ref{fig: criterions}).  As most eddies are not perfectly circular, we allow the angular difference to fall in the range [$-Sa$,0]. $Sa$ is arbitrarily set at 108° to allow for elongated elliptical flow.

Given the potential for turbulence and even flow artifacts, we anticipate that there will be occasions where a small kink in the flow (see Figure \ref{fig: criterions}, box 2a) results in a positive rather than negative angular difference.  Hence we allow a few small (< $Sae$) positive angular differences, with the maximum number of exceptions specified by $San$. Since our minimum verification starts with a radius of 3 voxels, which means 16 sample points along the test path, we could set the maximum positive angular difference ($Sae$) to 360/16 = 22.5°. Alternatively, a 10\% sector of the circle is 18°; we choose this later value for the results shown in section 5.

\begin{figure}[tb]
    \centering
    \includegraphics[width=0.8\columnwidth]{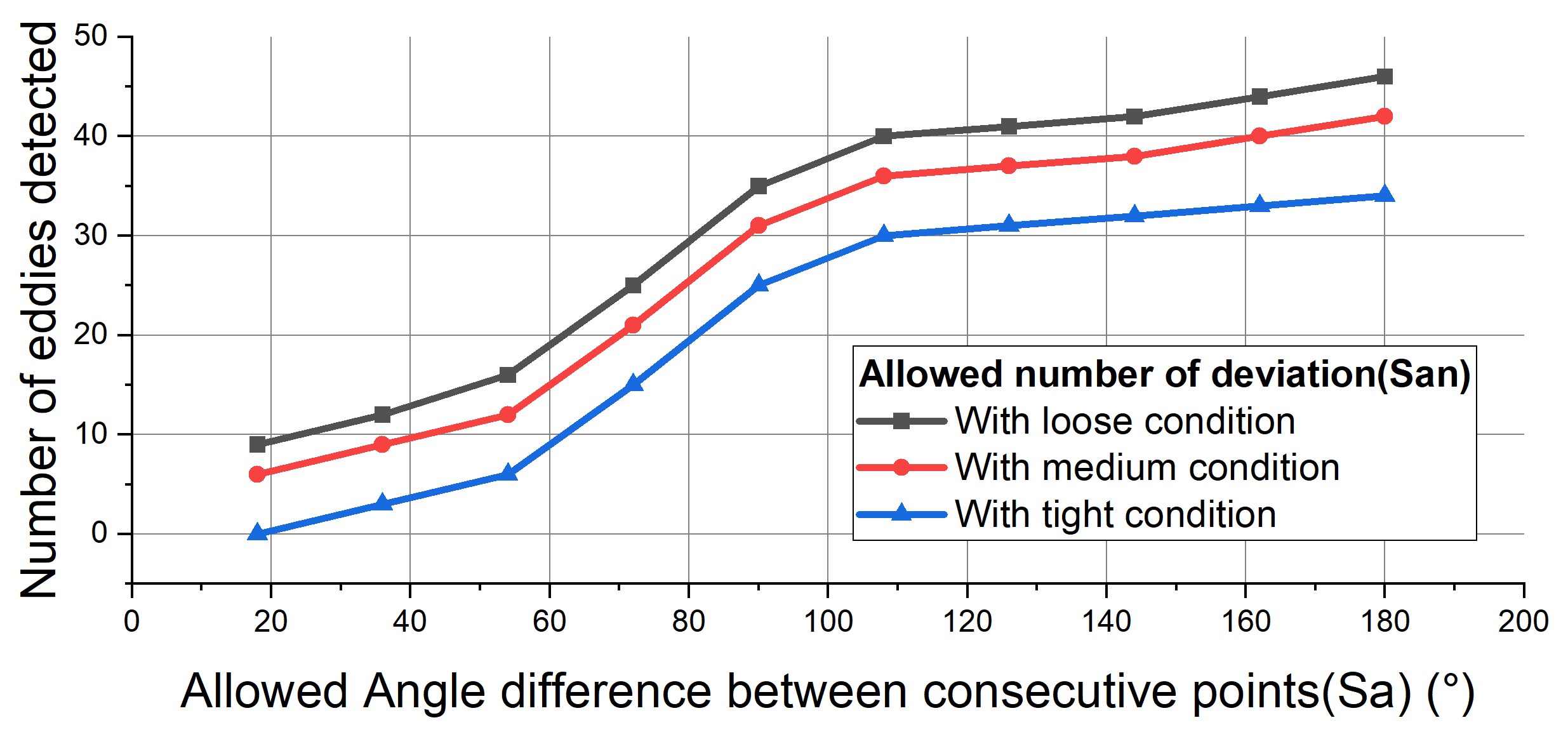}
    \caption{The detection result changed with $Sa$ and $San$. }
    \label{fig:Sa and San}
\end{figure}

Figure \ref{fig:Sa and San} shows how the number of eddies detected varies with with different $Sa$ and $San$ parameters. The number of detected centers flattens as the threshold for negative angular difference ($Sa$) is increased above 100°.  True eddy structures should not have large angular differences along their border but may not have a precise circular shape.  The loose condition allows significant deviations in angular difference towards small positive values while the tight condition strictly constrains the angular difference to small-to-moderate negative values.  Setting $Sa$ to 108° degrees and using a tight condition seems advantageous in retaining likely eddy candidates.

\subsection{Computation Time}\label{time}
We also record the computation time for three different approaches. Not surprisingly, the winding angle approach is computationally much more expensive than the other two approaches. Table \ref{tab:computation time} shows the computation time for different approaches. In our experiments, the winding angle approach is 10x slower than the other two approaches. On the other hand, several studies \cite{friederici2021winding} report much faster processing times.

\begin{table}
  \caption{Computation Time for different approaches}
  \label{tab:computation time}
  \scriptsize%
	\centering%
  \begin{tabu}{%
	r%
	*{7}{c}%
	*{2}{r}%
	}
  \toprule
   approach & {Winding Angle (only sea surface)} &   {OW Value} &   {Hybrid}\\
  \midrule
  Time (s) & 330.6640 & 26.3967 & 21.0859 \\
  \bottomrule
  \end{tabu}%
\end{table}

Compared with other approaches, this new hybrid method gives a faster and more effective method to detect the border of the eddy. However, we're also interested about underlying causes behind these results from different approaches. In the next section, we will discuss the potential explanation of these results and also apply this approach to more datasets.

\section{Discussion}

Now we compare a standard implementation of the winding angle approach and the previous OW thresholding results with our new results as a benchmark of our successes. We introduce our winding angle results, discuss the differences between the methods, and show applications to other datasets.

\subsection{Comparison with other approaches}\label{windingangle}
\begin{figure}[tb]
    \centering
    \includegraphics[width=0.9\linewidth]{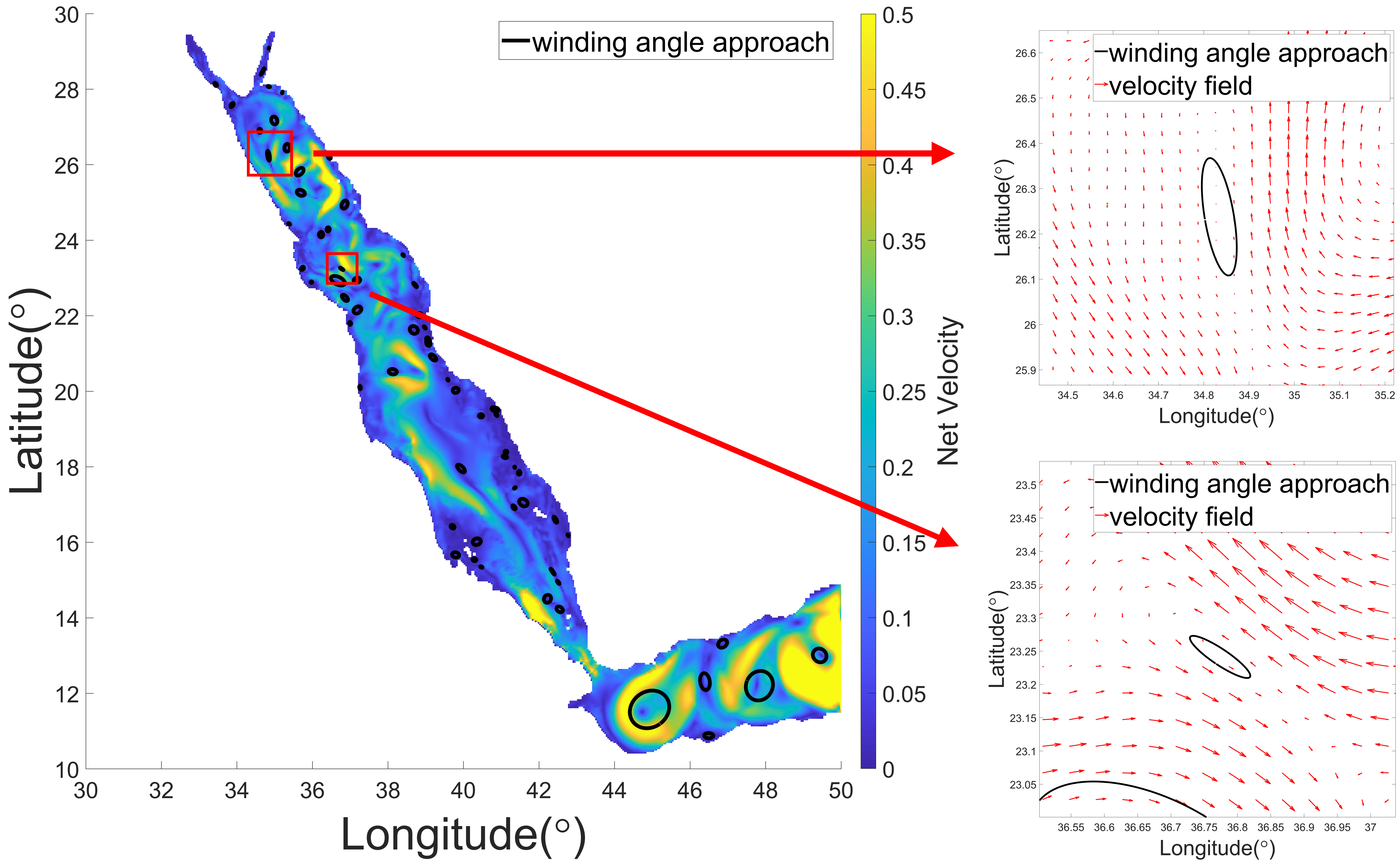}
    \caption{The visualization of the extraction result by using the winding angle based approach. The figure on the left visualizes the extraction result on the sea surface; Two figures on the right visualize two sub-region that are detected by the winding angle based approach as false eddy structures.}
    \label{fig: winding angle}
\end{figure}

We use a winding angle approach as a benchmark to evaluate our new hybrid approach. This approach detects 65 objects in the first frame of ensemble member 1 in the Red Sea dataset. To assess whether the additional 39 detections are false positives in the winding angle method or false negatives in our hybrid method, we look at the velocity fields close up. Many of the additional detections are small features in the northern Red Sea; visual inspection (as in Figure \ref{fig: winding angle}) suggests these rarely have any real swirl component but rather reflect complex changes in the background flow structure.  In contrast, the larger features identified in the Gulf of Aden show much greater conformance in eddy detection between methods.

\begin{figure}[tb]
    \centering
    \includegraphics[width=.99\columnwidth]{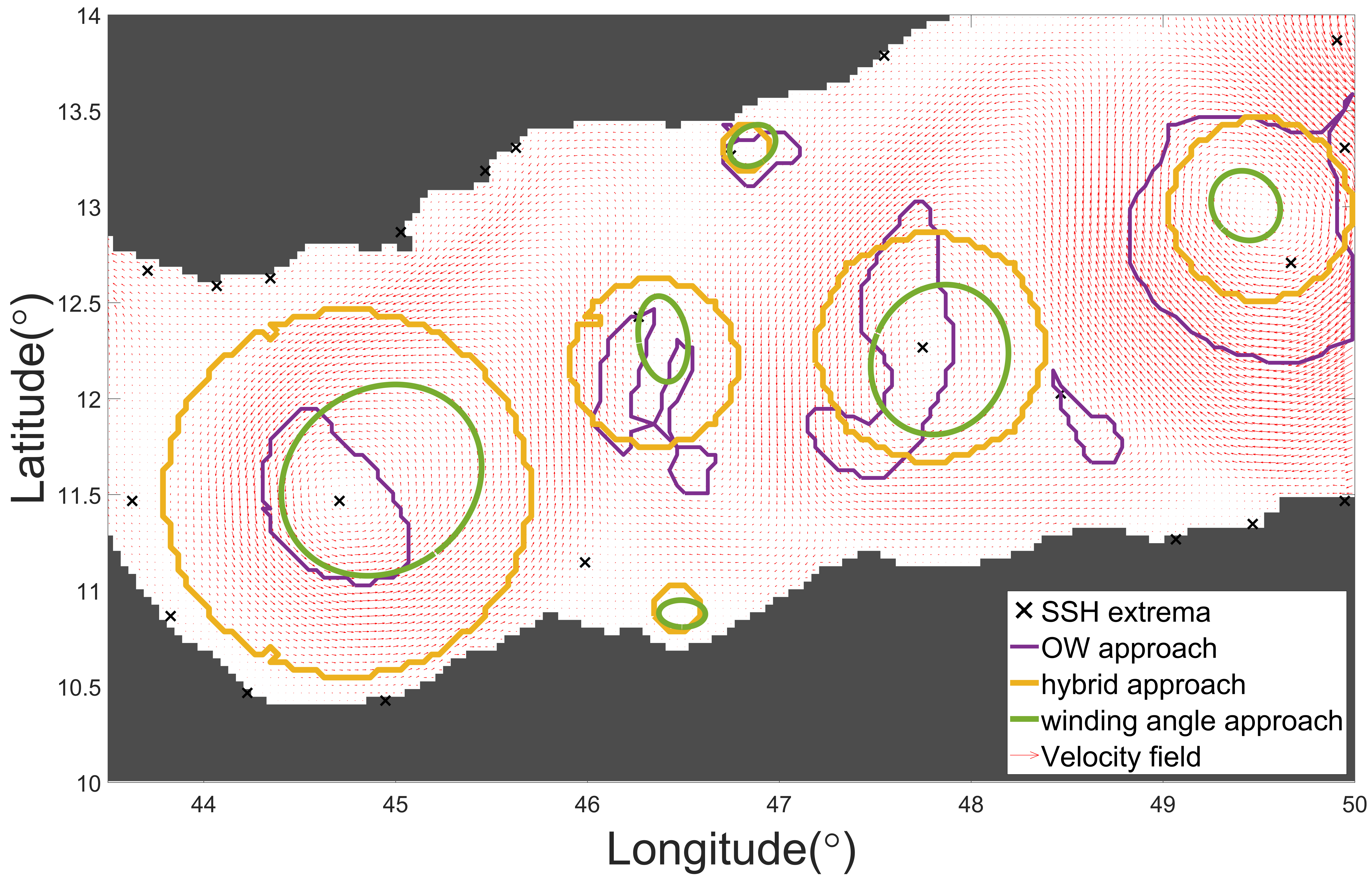}
    \caption{A comparison of multiple approches for the extracted border of detected eddies. The results are obtained by using the OW based approach, the hybrid approach and the winding angle approach at Gulf of Aden on the Red Sea dataset. }
    \label{fig: comparison between methods}
\end{figure}

\subsection{The difference of three approaches}
Figure \ref{fig: comparison between methods} compares detection results of the winding angle approach, the OW based approach and the new hybrid approach at the Gulf of Aden in the frame 1 of ensemble member 1 in Red Sea dataset. Although the OW-based approaches have been previously demonstrated to provide a universal and straightforward method to detect vortex (eddy) cores \cite{rave2021multifaceted}, here the OW-derived boundaries imply asymmetrical and fragmentary structures despite the accurate core detection. In contrast, both the winding angle approach and our new hybrid approach yield elliptical to circular extraction boundaries for the same eddy centers. Our new hybrid method extracts the broader region of eddy influence while the winding angle method often extracts only the core of the eddy. Looking deeper at the three methods, they represent the three kinds of eddy detection approach: value-based, geometry-based and hybrid. The OW parameter is  fast and intuitive. However, the OW parameter is strongly affected by the background flow in the dataset, which can result in detection of regions of abrupt change in flow regardless of the geometry structure of the velocity field. On the other hand, the geometry-based approach follows precisely the geometry of the velocity field. The winding angle approach detects eddy structures by clustering streamlines which may miss some features. Therefore, we can find in Figure \ref{fig: comparison between methods} that eddies detected by winding angle approach usually only contain the core region of the eddy. But adjustment of the parameters of the winding angle approaches might provide a boundary closer in size to our hybrid approach. 

Our hybrid approach combines the advantages of OW and winding angle approaches. We use a predefined path to check velocity components on the border, which yields a rough boundary of the eddy represents its footprint or region of influence rather than an exact boundary. The extracted structure enables us to obtain the overall 3D shape and size of the eddy footprint and extract the internal distribution of oceanographic properties such as temperature. While the predefined path limits the accuracy of 2D shape estimates, the rapid estimation of 3D volume and eddy depth penetration brings a comprehensive understanding of the transport potential of eddies. 

\subsection{Comparison with 2020 SciVis Contest results}

\begin{figure}[tb]
    \centering
    \includegraphics[width=\columnwidth]{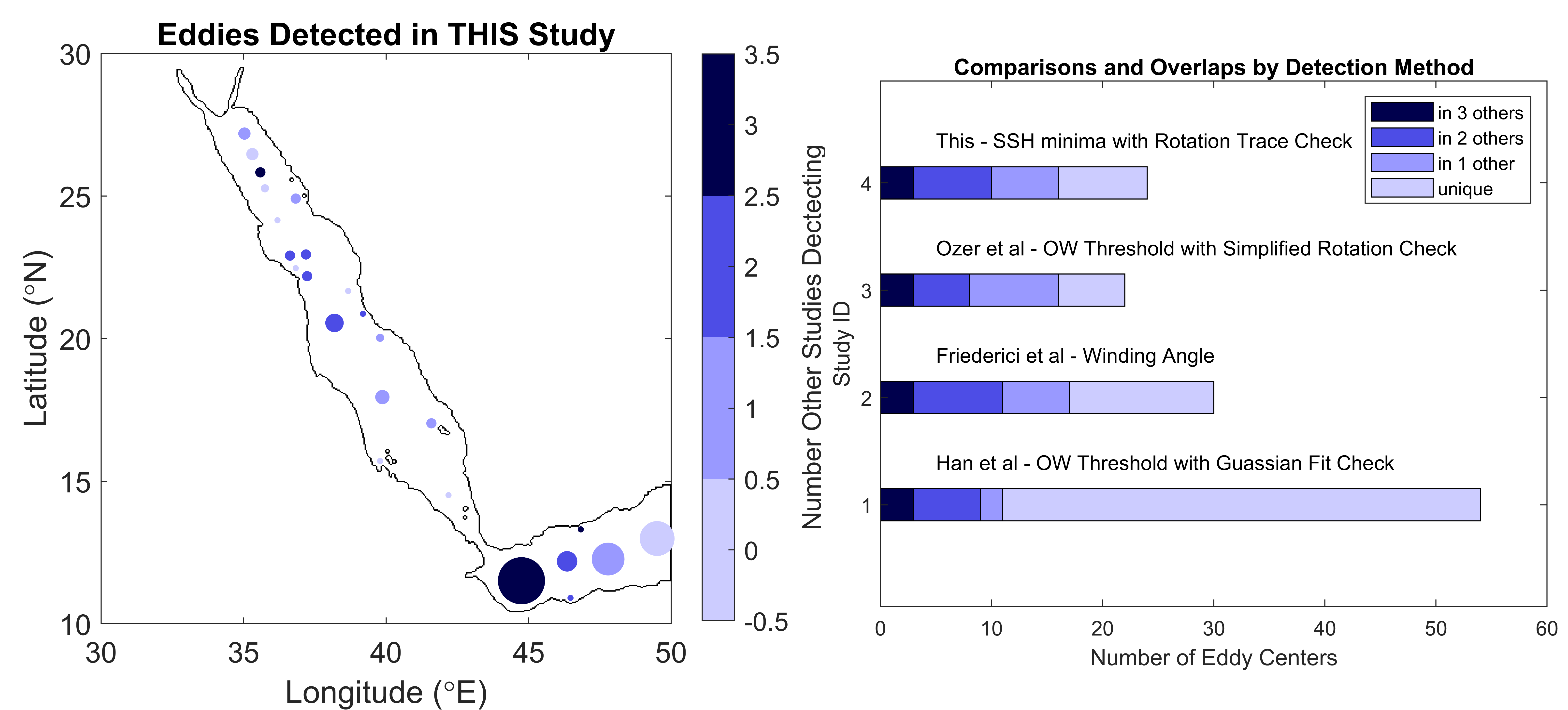}
    \caption{Comparison of our new approach and selected other approaches from Scivis Contest 2020 competitors.See listings in Scivis Contest (a) Map shows which eddies detected in the new approach were also detected in other approaches. (b) Bar chart documents the overlaps and uniqueness of detections across 4 studies.}
    \label{fig:comparison with other methods}
\end{figure}

Using figures from three of the SciViz Contest 2020 submissions, we compared where the other studies detected eddies with our results. Figure \ref{fig:comparison with other methods} above shows the locations of our detected eddies, indicating with color which ones are common across the methods. The methods vary in the number of centers detected with only 3 eddies out of over 50 possible centers common to all four included studies. Compared with the previous approaches, our new approach detects a greater number of structures in the Gulf of Aden (although the ensemble-averaging approach in \cite{rave2021multifaceted} also detected all of the same structures), but fewer, although more separated, structures in the northern Red Sea.  More broadly, every study detected a significant number of unique centers suggesting that eddy center detection is highly sensitive to the choice of thresholds and criteria.  In addition, the rotation checks clearly have the potential to eliminate a number of spurious candidates for all types of approaches.

\subsection{Results from more datasets}

 To test how particular our method is to the Red Sea dataset, we applied our hybrid method to additional datasets (Figure \ref{fig:more dataset}).  In addition to applying the basic extraction discussed above, we track the eddies over time and visualize their properties, such as velocity, temperature or salinity, along the eddy boundary and surface. Figure \ref{fig:more dataset} shows our detection results (first 50 detected eddies) for both a North Atlantic dataset and a North Pacific dataset\cite{metzger2017global}. The tracked path (in purple) ends at the final state of the eddy over this time period. Such tracked paths can potentially lead to new oceanographic insights about the nutrition, ocean flow or prediction.

\section{Conclusion}

This study investigated using the velocity field directly to extract eddy structures from the Red Sea, North Atlantic, and North Pacific ocean simulations. After re-analyzing several critical issues in previous OW-based eddy detection approaches, a new hybrid extraction approach was proposed. The new hybrid method detects eddy structures by combining SSH and velocity fields with geometric constraints to assess rotation coherence without calculating the velocity streamline. A comparison with previous approaches confirmed both the importance of rotation checks and the sensitivity of results to methodology, value thresholds, and criteria.  Examination of the 3D structure in comparison with typical patterns of eddy properties confirms that the structure of individual eddies is more complex and variable than the ideal or mean eddy.  

Future work will assess the reliability of extracting eddy center and outer structure in comparison with previous studies, e.g. \cite{matsuoka2016new}, and address the behavior of eddy structures across time and ensembles. Ensemble and time series averaging should confirm the typical structure is an average rather than instantaneous feature. Tracking of eddy movement and changes in eddy size and structure (including internal temporal coherence) can be pursued by harnessing our proposed detection methods to tracking methods. The input of domain scientists will be critical to analyzing the physical properties of the eddy, as we incorporate eddy tracking and auto-classification technology.

\cite{10.2312:envirvis.20231101}


\bibliographystyle{eg-alpha-doi}

\bibliography{ReferenceCitation}

\end{document}